\RequirePackage{fix-cm}
\documentclass[twocolumn]{svjour3}          
\smartqed  

\usepackage{cite}
\usepackage[pdftex]{graphicx}
\usepackage{ragged2e}
\usepackage[tight,footnotesize]{subfigure}
\usepackage{rotating}
\usepackage{graphicx}
\usepackage{amsmath,amssymb} 
\usepackage{array}
\usepackage{multirow}
\usepackage{colortbl}
\usepackage{amsfonts}
\usepackage{pifont}
\usepackage{xspace}
\usepackage{etoolbox}
\usepackage{overpic}
\usepackage{color}
\usepackage{microtype}
\usepackage{booktabs}

\newcommand{\figref}[1]{Fig. \ref{#1}}
\newcommand{\tabref}[1]{Tab. \ref{#1}}

\newcommand{\secref}[1]{Sec. \ref{#1}}

\newcommand{\orcid}[1]{\href{https://orcid.org/#1}{\includegraphics[width=10pt]{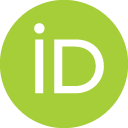}}}

\def\etc{{\em etc.}}
\def\etal{{\em et al}}

\usepackage{hyperref}
\hypersetup{breaklinks=true,citecolor=blue, colorlinks}

\graphicspath{{./figs/}}
\DeclareGraphicsExtensions{.pdf,.jpg,.png}

\usepackage{silence}
\journalname{Research Article}

\begin{document}

\title{PAGS: Autofocusing Photoacoustic Tomography via Speed-of-Sound-Adaptive Gaussian Splatting}

\titlerunning{PAGS for Autofocusing Photoacoustic Tomography}

\author{Jiarui Ge$^{1}$ \and
Jintao Ma$^{2}$ \and
Bangxu Fan$^{2}$ \and
Jinyan Zhang$^{2}$ \and
Xiaokang Yang$^{1}$ \and
Shuai Na$^{2,3,4,5,*}$\thanks{Corresponding author: \email{shuai@pku.edu.cn}.} \and
Xiaoyun Yuan$^{1,*}$\thanks{Corresponding author: \email{yuanxiaoyun@sjtu.edu.cn}.}
}

\authorrunning{J. Ge \etal}

\institute{
$^{1}$MoE Key Lab of Artificial Intelligence, Institute of AI, School of Computer Science, Shanghai Jiao Tong University, Shanghai, China.\\
$^{2}$National Biomedical Imaging Center, College of Future Technology, Peking University, Beijing, China.\\
$^{3}$Academy for Advanced Interdisciplinary Studies, Peking University, Beijing, China.\\
$^{4}$Key Laboratory of Machine Perception, Ministry of Education, Beijing, China.\\
$^{5}$PKU--Nanjing Institute of Translational Medicine, Nanjing, China.
}

\date{}

\maketitle

\begin{figure*}[t]
  \centering
  \includegraphics[width=\textwidth]{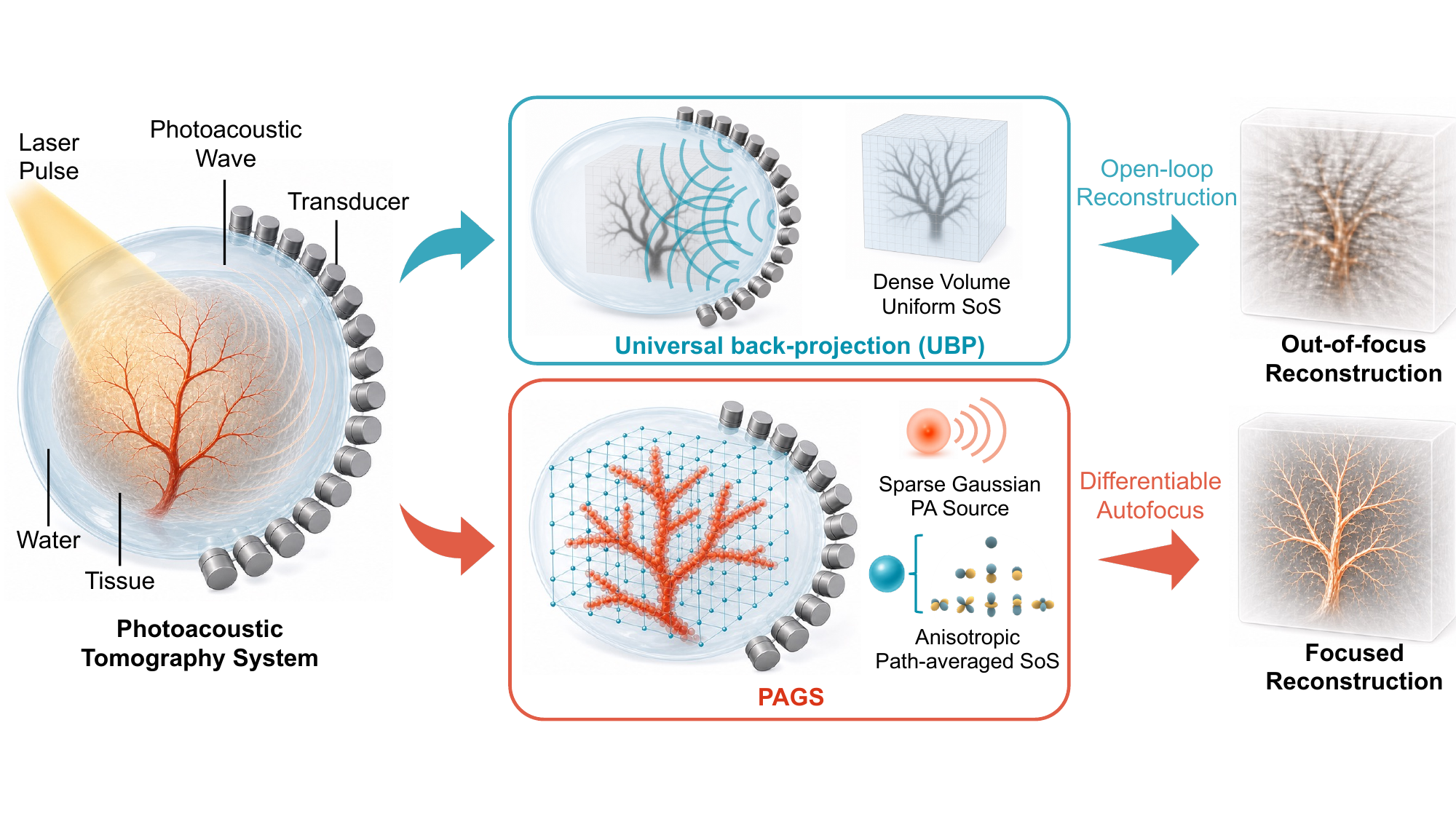}
  \caption{Overview of the proposed PAGS framework for blind autofocusing photoacoustic tomography. Compared with universal back-projection (UBP) under a uniform speed-of-sound (SoS) assumption, PAGS represents the initial pressure field with sparse Gaussian photoacoustic (PA) sources and learns an anisotropic path-averaged SoS (ASoS) field with spherical harmonic (SH) probes, enabling focused reconstruction without calibrated SoS priors.}
  \label{fig:teaser}
\end{figure*}

\begin{abstract}
Photoacoustic computed tomography (PACT) combines optical absorption contrast with acoustic detection for high-resolution deep-tissue imaging. A persistent challenge is that unknown speed-of-sound (SoS) heterogeneity changes acoustic time-of-flight, causing defocusing artifacts when reconstruction assumes a uniform SoS. Existing SoS-adaptive methods either rely on calibrated acoustic priors or optimize dense physical medium models, which becomes expensive and difficult to scale in 3D. We propose PAGS, a differentiable framework for blind autofocusing PACT via speed-of-sound-adaptive Gaussian splatting. PAGS represents the initial pressure field with sparse Gaussian photoacoustic (PA) sources and replaces explicit medium recovery with a compact anisotropic path-averaged SoS (ASoS) field parameterized by spherical harmonic probes. This latent propagation field directly controls source-to-transducer arrival-time alignment, while an analytic Gaussian acoustic projection maps the source representation to transducer signals efficiently. The resulting closed-loop signal-domain optimization jointly updates the Gaussian PA source parameters and the ASoS field from measured data, without calibrated SoS priors. Experiments on simulated and physical phantom data demonstrate improved reconstruction sharpness under heterogeneous acoustic media, robustness to sparse-view sampling, and computational benefits from the analytic Gaussian projection.



\end{abstract}

\ESM{Supplementary Video~1 provides rotating three-dimensional comparisons of vanilla UBP, Dual-SoS UBP, SlingBAG, and PAGS, together with PAGS reconstructions at different sampling ratios.}

\section{Introduction}
\label{sec:intr}

Photoacoustic computed tomography (PACT) is a non-invasive imaging modality that combines optical absorption contrast with acoustic detection~\cite{lin2025survey}. As illustrated in Fig.~\ref{fig:teaser}, a short laser pulse excites light-absorbing chromophores, such as hemoglobin or melanin, and the resulting thermoelastic expansion generates broadband ultrasonic waves that are recorded by surrounding transducers. Reconstructing the initial pressure field from these time-resolved acoustic measurements enables high-resolution structural and functional mapping of biological tissues. This capability has made PACT useful in a wide range of preclinical and clinical studies, including vascular imaging, neuroimaging, and tumor characterization.

The quality of PACT reconstruction, however, is strongly affected by the assumed speed of sound (SoS). Standard analytical reconstruction methods, such as universal back-projection (UBP)~\cite{xu2005universal}, commonly assume a globally uniform SoS. Real tissue and surrounding coupling media often have different acoustic speeds, so the same photoacoustic source can have different path-averaged SoS values toward different transducers~\cite{poimala2024compensating,lin2025survey}. These path-dependent arrival-time errors lead to defocusing, geometric distortion, and loss of fine vascular detail. Correcting unknown SoS-induced aberration is therefore a central problem for high-resolution PACT.

Existing SoS-adaptive reconstruction strategies face a difficult trade-off. Prior-informed multi-SoS reconstruction can improve focusing when tissue boundaries and regional SoS values are known, but such calibrated acoustic priors are often unavailable and may require additional segmentation or auxiliary measurements. More general iterative or learning-based approaches can optimize a dense physical medium or scalar SoS map jointly with the absorption distribution~\cite{poimala2024compensating,li2025coordinate}, but the arrival-time correction needed for focusing is obtained only after path integration through that medium. This high-dimensional parameterization and indirect forward dependency make blind 3D optimization computationally expensive and prone to poor local solutions.

We propose PAGS, a differentiable framework for blind autofocusing PACT via speed-of-sound-adaptive Gaussian splatting. Building on Gaussian-based reconstruction~\cite{kerbl20233d,li2024slingbag}, the key idea is to replace explicit recovery of the full acoustic medium with a latent anisotropic path-averaged SoS (ASoS) field that directly controls source-to-transducer arrival-time alignment. PAGS represents the initial pressure field with sparse Gaussian photoacoustic (PA) sources and represents the propagation effect with compact spherical harmonic (SH) probes anchored on a spatial grid. For each Gaussian PA source and transducer direction, PAGS queries the ASoS field, projects the source to the transducer signal using an analytic Gaussian acoustic model, and optimizes the source and propagation parameters in a single signal-domain loop.

This design aligns the representation with the structure of the inverse problem. The PA source distribution is sparse and localized, making Gaussian PA sources a natural continuous primitive for vascular structures. The heterogeneous acoustic medium is not recovered directly; instead, its accumulated effect on propagation is expressed as a smoother, lower-dimensional ASoS field over source locations and transducer directions. As a result, PAGS focuses on the propagation variable that directly affects reconstruction sharpness while avoiding the cost and ambiguity of dense medium recovery.

The main contributions are summarized as follows:
\begin{enumerate}
    \item We introduce a latent ASoS propagation field for blind PACT autofocusing. It models direction-varying path-averaged SoS directly, avoiding explicit dense acoustic medium recovery and calibrated SoS priors.
    \item We couple the ASoS field with sparse Gaussian PA sources and an analytic Gaussian acoustic projection, enabling efficient differentiable optimization from measured transducer signals.
    \item We validate PAGS on simulated and physical phantom data, including sparse-view settings and component ablations, showing improved focusing under heterogeneous acoustic media and favorable efficiency from the analytic projection.
\end{enumerate}

\section{Related Work}
\label{sec:related}

\subsection{PACT Reconstruction Methods}
Conventional PACT reconstruction algorithms are broadly categorized into analytical and iterative paradigms. Analytical methods, predominantly Universal Back‑Projection (UBP)~\cite{xu2005universal} and Delay‑and‑Sum (DAS), are widely adopted for their computational efficiency. By deterministically mapping time‑resolved signals to the spatial domain, they operate on the highly idealized assumption of straight‑ray propagation within a homogeneous medium. However, this assumption critically fails in realistic biological tissues~\cite{lin2025survey,poimala2024compensating}. Acoustic heterogeneities across diverse tissue layers induce wavefront refraction and significant time‑of‑flight (ToF) mismatches. Since manual global SoS tuning cannot compensate for these localized phase errors, analytical reconstructions inherently suffer from geometric distortions, spatial blurring, and severe resolution degradation~\cite{yang2025skull}. Furthermore, these methods strictly rely on dense spatial sampling to guarantee reconstruction fidelity. Under sparse sensor coverage or limited‑view geometries, they lack the inherent mechanisms to constrain the solution space, inevitably generating streak artifacts and making reliable structure identification exceedingly difficult~\cite{li2024slingbag,lin2025survey}.

To address these limitations, model‑based iterative reconstruction methods treat PACT as an inverse problem under an optimization framework~\cite{huang2013full,wang2022learned,liu2024accelerated}. By integrating regularization terms such as Total Variation (TV) or $L_1$‑norm penalties, these frameworks achieve superior resilience against sparse sampling and measurement noise. For example, Huang \textit{et al.}~\cite{huang2013full} developed a full‑wave iterative reconstruction framework that explicitly accounts for acoustic heterogeneity, demonstrating that incorporating true SoS maps significantly improves image fidelity in realistic tissue models. More recent works further introduce learned regularization priors~\cite{wang2022learned} and multi‑channel autoencoder constraints~\cite{liu2024accelerated} to accelerate convergence and enhance robustness. Time‑reversal approaches~\cite{kowar2014time} also exploit the symmetry of the wave equation for efficient reconstruction, though they are still sensitive to SoS mismatches. However, the repeated evaluation of the forward acoustic propagation and its adjoint gradient inherently incurs substantial computational and memory costs. Furthermore, standard iterative frameworks typically rely on a predefined SoS map for accurate forward modeling. In realistic biological tissues, acquiring the exact anatomical SoS distribution is highly challenging. Inaccuracies in this static prior can compromise the compensation of acoustic aberrations, ultimately limiting the overall reconstruction fidelity~\cite{song2023sparse}.

In recent years, deep learning has emerged as a powerful alternative for PACT reconstruction. Data‑driven architectures, ranging from U‑Nets to diffusion models, leverage rich learned priors to achieve striking artifact reduction under ill‑posed conditions, such as sparse‑view or limited‑aperture scenarios~\cite{song2023sparse,liu2024recent,lin2025survey,antholzer2019deep}. While these purely data‑driven networks offer impressive structural restoration, pioneering physics‑aware frameworks have further attempted to explicitly incorporate physical constraints to dynamically resolve unknown acoustic aberrations. For instance, Hauptmann \textit{et al.}~\cite{hauptmann2018model} proposed a model‑based learning scheme that couples an iterative inversion with a learned regularization, achieving high‑quality limited‑view 3D reconstructions. Poimala \textit{et al.} designed a specialized network to compensate for unknown SoS distributions, effectively recovering spatial resolution degraded by wavefront aberrations~\cite{poimala2024compensating}. Similarly, Li \textit{et al.} employed coordinate‑based implicit neural representations to directly recover a spatial SoS field jointly with the absorption distribution, enabling aberration correction through physically informed decoding~\cite{li2025coordinate}. Together, these learning‑based and physics‑aware approaches demonstrate the immense potential of integrating data‑driven priors with acoustic physical models. However, despite their success in accounting for non‑uniform SoS, representing the acoustic field via dense grids or implicit functions remains computationally demanding, limiting their scalability and memory efficiency in complex 3D environments.

\subsection{Gaussian‑based Reconstruction Methods}

3D Gaussian Splatting (3DGS)~\cite{kerbl20233d,yu2024mipsplatting,lu20242dgs,liang2024scaffoldgs} has emerged as a powerful explicit scene representation paradigm. Instead of relying on traditional voxels or meshes, it parameterizes a scene using a collection of learnable, anisotropic 3D Gaussian primitives. Each primitive is explicitly defined by its position, covariance, opacity, and view‑dependent color. Rendered via a highly efficient differentiable tile‑based rasterizer, 3DGS achieves real‑time performance while preserving state‑of‑the‑art visual fidelity. Recent extensions have further improved anti‑aliasing~\cite{yu2024mipsplatting}, geometric accuracy~\cite{lu20242dgs}, and structured view‑adaptive rendering~\cite{liang2024scaffoldgs}, solidifying 3DGS as a robust spatial modeling framework.
This structural flexibility and computational efficiency have inspired a surge of dynamic extensions. For instance, Deformable 3DGS~\cite{yang2024deformable} introduces a per‑Gaussian temporal deformation field, allowing the framework to reconstruct high‑quality geometry and appearance from monocular dynamic videos. Building upon the concept of temporal modeling, 4DGS~\cite{wu20244d} directly embeds the optimization into the space‑time domain, jointly updating Gaussian attributes and motion trajectories to enable real‑time dynamic view synthesis. Furthermore, Spacetime Gaussian Feature Splatting~\cite{li2024spacetime} explicitly models the evolution of both geometry and appearance using polynomial trajectories, which significantly enhances the temporal coherence of dynamic renderings. 
Collectively, these advances highlight the core strengths of Gaussian‑based representations. As explicit, grid‑free primitives, they can efficiently capture fine geometric details, seamlessly adapt to complex topological changes, and maintain full end‑to‑end differentiability, establishing 3DGS as a highly expressive and robust spatial modeling framework.

Building upon this momentum, Gaussian representations have been rapidly adapted to medical tomographic reconstruction. In X‑ray computed tomography (CT), $R^2$-Gaussian~\cite{zha2024r2gaussian} introduced a radiometrically corrected Gaussian splatting forward model that compensates for scattering and beam hardening effects, achieving high‑quality 3D reconstructions under sparse‑view acquisition. Similarly, for 3D ultrasound volume imaging, UltraGauss~\cite{eid2025ultragauss} integrated learnable Gaussian primitives with an acoustic wave propagation renderer to build a fast reconstruction framework, demonstrating strong adaptability to reflection and attenuation physics. In whole-body positron emission tomography (PET), GR-Diffusion~\cite{geng2026grdiffusion} combines Gaussian representations with diffusion models to improve low-dose reconstruction. MedGS~\cite{marzol2025medgs} extends Gaussian splatting to unified multi-modal 3D medical imaging. These successful cross-modal adaptations confirm that, when equipped with modality-specific differentiable forward models, Gaussian primitives can serve as a highly expressive and versatile basis for solving complex volumetric inverse problems.

In the realm of PACT, SlingBAG~\cite{li2024slingbag} recently pioneered the integration of Gaussian representations by parameterizing the initial photoacoustic pressure as a collection of learnable 3D Gaussian PA sources. For acoustic forward projection, it approximates each continuous Gaussian primitive by multiple concentric spherical shells and uses a smoothed sphere response with error‑function evaluations to maintain differentiability. While effective, this approximation increases the per‑iteration computational cost and still fundamentally hinges on a globally homogeneous SoS assumption. Under this constraint, the acoustic time‑of‑flight from each Gaussian PA source to each detector is evaluated solely from Euclidean distance, neglecting path‑dependent aberrations caused by heterogeneous acoustic media. Consequently, the uniform‑SoS limitation leads to structural defocusing and geometric distortion, compromising spatial resolution and reconstruction fidelity.

These limitations motivate PAGS, an SoS‑adaptive Gaussian splatting framework for blind autofocusing PACT. PAGS keeps the sparse Gaussian PA source representation, replaces the multi‑sphere approximation with an analytic Gaussian acoustic projection, and introduces a learnable anisotropic path‑averaged SoS field parameterized by compact spherical harmonic (SH) probes. In this way, PAGS optimizes the source representation and the path‑level propagation correction directly from measured transducer signals, without recovering a dense physical SoS map.

\section{Method}

\begin{figure*}[t]
    \centering
    \includegraphics[width=\textwidth]{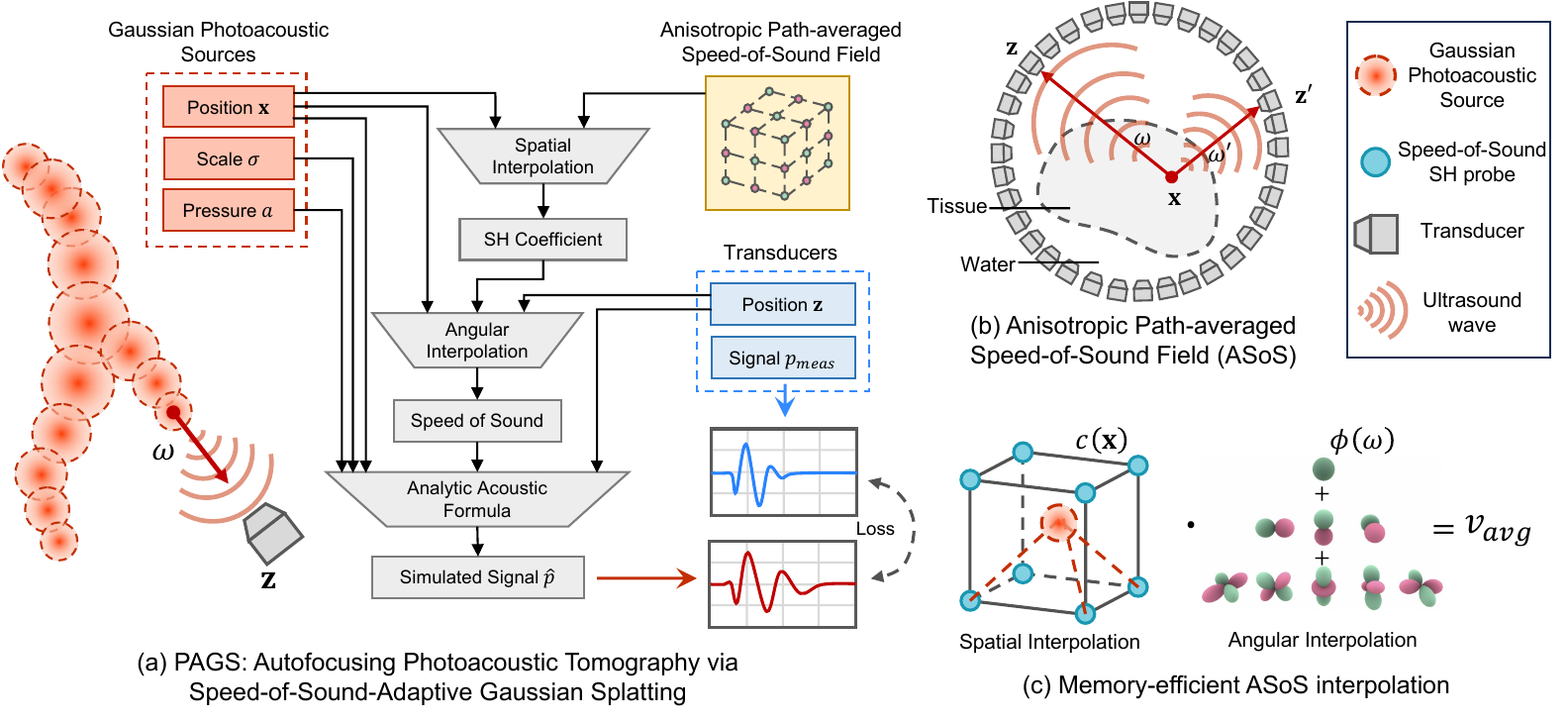}
    \caption{Core implementation details of PAGS. (a) The differentiable reconstruction model couples Gaussian photoacoustic (PA) sources, anisotropic path-averaged SoS coefficients, and transducer measurements through an analytic acoustic forward model. (b) The anisotropic path-averaged SoS (ASoS) field represents the direction-varying path-averaged SoS from each source location to the transducer array, implicitly capturing the accumulated propagation effect of heterogeneous media. (c) The ASoS query uses two interpolations: spherical harmonic (SH) coefficients are first obtained at the PA source position by spatial interpolation from neighboring probes, and the path-averaged SoS is then obtained by angular SH interpolation along each source--transducer direction.}
    \label{fig:method_details}
\end{figure*}

\subsection{Preliminaries}
\label{sec:preliminaries}

PAGS builds on the sparse source representation introduced by SlingBAG~\cite{li2024slingbag}. Given time-resolved acoustic measurements recorded by an array of transducers, the reconstruction target is the initial pressure field $p_0(\mathbf{r})$, where $\mathbf{r}$ is a 3D coordinate in the imaging volume. Following SlingBAG, we represent this field as a set of Gaussian photoacoustic (PA) sources:
\begin{equation}
p_0(\mathbf{r})
=
\sum_{i=1}^{N}
a_i
\exp\left(
-\frac{\lVert \mathbf{r}-\mathbf{x}_i\rVert^2}{2\sigma_i^2}
\right),
\end{equation}
where $N$ is the number of Gaussian PA sources, and $\mathbf{x}_i$, $a_i$, and $\sigma_i$ denote the center, pressure amplitude, and spatial scale of the $i$-th source. This representation is compact, continuous, and differentiable, making it well suited for localized PA structures such as vessels.

However, directly reusing this Gaussian PA source backbone leaves heterogeneous acoustic propagation unaddressed. SlingBAG projects Gaussian PA sources to transducer signals under a globally uniform speed of sound (SoS), so its predicted time-of-flight depends only on source--transducer distance. When the actual propagation is heterogeneous, this uniform-SoS assumption introduces arrival-time errors and leads to out-of-focus reconstruction. PAGS keeps the compact Gaussian PA source representation, but redesigns the propagation and forward-projection modules described below.

\subsection{Framework Overview}
\label{sec:framework_overview}

PAGS formulates blind autofocusing as a closed-loop differentiable signal-matching problem. As illustrated in Fig.~\ref{fig:method_details}(a), its learnable variables are the Gaussian PA source parameters and the propagation parameters that map these sources to the measured transducer signals. A forward pass starts from the current Gaussian PA sources, queries the anisotropic path-averaged SoS (ASoS) field for each source--transducer pair, projects each source with an analytic acoustic model, and superposes all source contributions to synthesize the detector signals.

The signal residual then updates both parts of the representation. The Gaussian PA source parameters refine the reconstructed initial pressure field, while the ASoS field corrects the arrival-time alignment that controls focusing. Thus, PAGS differs from open-loop reconstruction by allowing source reconstruction and propagation correction to refine each other within the same signal-domain objective.

The central contribution of PAGS is to replace explicit acoustic medium recovery with a latent ASoS propagation field for autofocusing. Existing SoS-adaptive methods often optimize a dense medium or scalar SoS map, and the arrival-time correction needed for reconstruction is obtained only after path integration through that map. This creates a high-dimensional parameterization with an indirect forward dependency: a local change in the medium affects image focus only after being accumulated along many source--transducer paths. PAGS instead learns the anisotropic path-averaged SoS between source locations and transducer directions. Each ASoS value already represents the accumulated propagation effect along one path, yielding a smoother and lower-dimensional optimization target whose value directly changes the predicted arrival time. Coupled with the analytic Gaussian acoustic projection, this formulation lets signal residuals jointly update the Gaussian PA source parameters and the ASoS field through a simpler differentiable reconstruction model.

The following subsections describe the main components in order: Sec.~\ref{sec:sos_field} defines the ASoS field and its SH probe parameterization, Sec.~\ref{sec:analytical} derives the analytic acoustic projection of Gaussian PA sources, and Sec.~\ref{sec:joint_opt} presents the joint optimization objective.

\subsection{Anisotropic Path-Averaged SoS Field}
\label{sec:sos_field}

We now define the ASoS field used by the propagation component. The ASoS field is a latent propagation representation whose queried value serves as the effective path-averaged SoS for converting a source--transducer distance into its acoustic delay. It is optimized to compensate for propagation-induced arrival-time errors in autofocusing, rather than to recover the local physical SoS distribution of the acoustic medium.

For a Gaussian PA source at $\mathbf{x}$ and a transducer at $\mathbf{z}$, let
\begin{equation}
R=\lVert\mathbf{z}-\mathbf{x}\rVert,\qquad
\boldsymbol{\omega}=\frac{\mathbf{z}-\mathbf{x}}{R}
\end{equation}
denote the source--transducer distance and propagation direction. The corresponding time-of-flight (ToF) is modeled as
\begin{equation}
\mathrm{ToF}(\mathbf{x}, \mathbf{z}) =
\frac{R}
{v_{\mathrm{avg}}(\mathbf{x}, \boldsymbol{\omega})},
\end{equation}
where $v_{\mathrm{avg}}(\mathbf{x},\boldsymbol{\omega})$ is the effective path-averaged SoS from source position $\mathbf{x}$ along direction $\boldsymbol{\omega}$.

Here, ``anisotropic'' describes direction variation in the path-averaged SoS observed from a source location. As illustrated in Fig.~\ref{fig:method_details}(b), rays from the same source can cross different proportions of water and tissue before reaching different transducers; their path-averaged SoS values therefore vary with direction.

To make this continuous field compact and optimizable, PAGS stores it on a regular 3D grid of SoS SH probes, rather than attaching SoS parameters to individual Gaussian PA sources. This decouples the propagation model from the source set: Gaussian PA sources can move, split, and be pruned during optimization, while the ASoS field remains anchored in space and varies smoothly. Each probe represents the angular profile of the path-averaged SoS using second-order spherical harmonics (SH). We collect the nine real SH basis functions evaluated at direction $\boldsymbol{\omega}$ as
\begin{equation}
    \boldsymbol{\phi}_{2}(\boldsymbol{\omega}) =
    \left[
    \phi_0(\boldsymbol{\omega}), \ldots,
    \phi_8(\boldsymbol{\omega})
    \right]^\top
    \in \mathbb{R}^{9}.
    \label{eq:sh_basis}
\end{equation}
The first basis function is a constant term, the next three basis functions capture first-order directional variation, and the remaining five basis functions capture second-order angular variation. An SoS SH probe stores the corresponding weights $\mathbf{c}=[c_0,\ldots,c_8]^\top\in\mathbb{R}^{9}$; evaluating the weighted sum of these bases gives the path-averaged SoS for a queried direction. 

At projection time, PAGS queries the ASoS value for each active Gaussian PA source and transducer direction. As shown in Fig.~\ref{fig:method_details}(c), the query has two steps. First, PAGS gathers the eight neighboring probes of the enclosing voxel and performs trilinear interpolation in weight space:
\begin{equation}
\mathbf{c}(\mathbf{x}) =
\sum_{i,j,k \in \{0,1\}}
\mathbf{c}_{i,j,k}\,\alpha_{i,j,k}(\mathbf{x}),
\label{eq:sos_spatial_interp}
\end{equation}
where $\alpha_{i,j,k}(\mathbf{x})$ are the standard trilinear interpolation weights. This gives a local SH weight vector at the Gaussian PA source position. Second, PAGS evaluates the local SH expansion along the source--transducer direction:
\begin{equation}
v_{\mathrm{avg}}(\mathbf{x}, \boldsymbol{\omega}) =
\mathbf{c}(\mathbf{x})^\top \boldsymbol{\phi}_{2}(\boldsymbol{\omega}).
\label{eq:sos_angular_eval}
\end{equation}
Equations~\eqref{eq:sos_spatial_interp} and~\eqref{eq:sos_angular_eval} therefore implement two interpolations: the former spatially interpolates the SH weights to the Gaussian PA source position, and the latter angularly evaluates the weighted SH bases along a transducer direction. During optimization, the queried ASoS is constrained to a fixed physically valid interval $[v_{\min},v_{\max}]$ to prevent nonphysical acoustic delays.

An additional benefit of this representation is memory efficiency. Consider a representative setting with 100K active Gaussian PA sources and 4.6K transducers. \textbf{(1) Learnable parameters.} A direct source-to-transducer ASoS table would cost 100K$\times$4.6K = 460M learnable scalars. PAGS instead optimizes the probe grid: $16^3$ probes with 9 SH weights per probe require only 36.9K learnable weights, a 12.5K$\times$ reduction. \textbf{(2) Query cache.} If angular evaluation were performed first, the query would immediately materialize one ASoS scalar per source-to-transducer pair, costing another 460M intermediate values. PAGS performs spatial interpolation first, producing only one temporary 9D weight vector per active source, i.e., 100K$\times$9 = 900K values, and evaluates the final source-to-transducer ASoS values on demand during signal synthesis. This reduces the intermediate query cache by about 510$\times$, from 460M to 900K values. The queried $v_{\mathrm{avg}}$ is then passed to the analytic acoustic forward model described next.

\subsection{Analytic Acoustic Forward Model}
\label{sec:analytical}

Given the path-averaged SoS queried from the ASoS field, PAGS next maps each Gaussian PA source to the transducer signal domain. SlingBAG maintains differentiability by approximating a Gaussian PA source with multiple smoothed spherical shells, but this approximation introduces repeated error-function evaluations and increases per-iteration cost~\cite{li2024slingbag}. PAGS instead uses the continuous Gaussian PA source directly and evaluates its acoustic response in closed form.

For a single Gaussian PA source with amplitude $a$ and spatial scale $\sigma$, the closed-form solution in an ideal uniform medium is
\begin{equation}
\begin{aligned}
p(R, t) &= \frac{a}{2R} (R - v_s t)
\exp\left( -\frac{(R - v_s t)^2}{2\sigma^2} \right) \\
&+ \frac{a}{2R} (R + v_s t)
\exp\left( -\frac{(R + v_s t)^2}{2\sigma^2} \right),
\end{aligned}
\end{equation}
where $v_s$ is the constant SoS and $R$ is the observation distance.

The first term is the outgoing wave and dominates at the detector. The second term corresponds to an incoming component whose envelope is $\exp(-2R^2/\sigma^2)$ near the detection time $t\approx R/v_s$; it is negligible in the far-field regime of PACT, where $R\gg\sigma$. Under the straight-path ASoS approximation, we therefore keep the outgoing term and replace the constant $v_s$ with the queried path-averaged SoS:
\begin{equation}
p(R, t; v_{\mathrm{avg}}) \approx
\frac{a (R - v_{\mathrm{avg}}t)}{2R}
\exp\left( -\frac{(R - v_{\mathrm{avg}}t)^2}{2\sigma^2} \right).
\end{equation}
This substitution treats the queried ASoS as an effective path-averaged SoS for the source--transducer path, so the analytic signal receives the learned arrival-time correction while retaining a closed-form differentiable response.

Compared with the multi-sphere approximation, this expression uses a single exponential evaluation for each source, transducer, and time sample. It also remains differentiable with respect to source position, amplitude, scale, and the ASoS coefficients through $R$ and $v_{\mathrm{avg}}$. Thus, each Gaussian PA source serves as both a compact image-domain primitive and an efficient unit for signal-domain optimization.

\subsection{Joint Optimization}
\label{sec:joint_opt}

We now assemble the ASoS query and analytic acoustic projection into the signal-domain objective. For the $i$-th Gaussian PA source and the $m$-th transducer, let
$R_{i,m}=\lVert\mathbf{z}_m-\mathbf{x}_i\rVert$ and
\begin{equation}
\bar{v}_{i,m}
=
v_{\mathrm{avg}}\left(
\mathbf{x}_i,
\frac{\mathbf{z}_m-\mathbf{x}_i}{R_{i,m}}
\right).
\end{equation}
Using $\bar{v}_{i,m}$ in the source-adaptive response, the synthetic pressure signal at transducer $\mathbf{z}_m$ is
\begin{equation}
\begin{aligned}
\hat{p}(\mathbf{z}_m, t)
&= \sum_{i=1}^{N}
\frac{a_i \Delta_{i,m}(t)}{2R_{i,m}}
\exp\left( -\frac{\Delta_{i,m}^2(t)}{2\sigma_i^2} \right),
\end{aligned}
\end{equation}
where $\Delta_{i,m}(t)=R_{i,m}-\bar{v}_{i,m}t$.

Let $p_{\mathrm{meas}}(\mathbf{z}_m,t)$ denote the measured pressure signal at the $m$-th transducer and time $t$. PAGS optimizes the Gaussian PA source parameters and the ASoS probe coefficients by minimizing the signal-domain discrepancy:
\begin{equation}
\mathcal{L}
=
\frac{1}{M T}
\sum_{m=1}^{M}
\sum_{t=1}^{T}
\left(
\hat{p}(\mathbf{z}_m,t)
-
p_{\mathrm{meas}}(\mathbf{z}_m,t)
\right)^2,
\end{equation}
where $M$ is the number of transducers and $T$ is the number of temporal samples.

Rather than introducing an explicit geometry-specific regularizer, PAGS controls the complexity of the ASoS field through its parameterization. Trilinear interpolation makes the queried field spatially continuous between probes, while the second-order SH basis limits angular variation to low-order modes. Thus, $\mathcal{L}$ is the complete optimization objective, with no assumption of a two-region medium or prescribed tissue--water interface.

Optimization uses Adam over the full computation graph. The learnable variables include the Gaussian PA source parameters $\{\mathbf{x}_i,a_i,\sigma_i\}$ and the SH probe coefficients $\mathbf{c}$ of the ASoS field. For source adaptation, we follow the density-control strategy of SlingBAG, splitting high-gradient sources and pruning negligible ones so that the sparse source set evolves with the reconstructed structure. A bandpass filter can optionally be applied to measured signals before optimization when out-of-band system noise is significant.

\section{Experiments}
\label{sec:experiments}

\subsection{Experimental Setup}
\label{subsec:exp_setup}

\subsubsection{Datasets}
We evaluate PAGS on two complementary datasets. The simulated dataset provides controlled geometry and a digital vascular phantom, while quantitative evaluation uses a reference reconstruction that incorporates realistic opto-acoustic and transducer responses. The physical phantom dataset tests robustness under real acquisition imperfections.

For the \textit{in silico} dataset, we simulate acoustic measurements from a 3D vascular phantom using the $k$-space pseudospectral method implemented in k-Wave~\cite{treeby2010k}. Fig.~\ref{fig:insilico_results} shows the simulation configuration, reconstruction results, and corresponding digital phantom. The left panels display transducer positions as purple-to-yellow gradient dots (top) and the acquisition geometry with a grey spherical cap, a yellow ellipsoid SoS interface, and red phantom voxels of 0.1~mm edge length (bottom). The remaining columns compare three orthogonal views of the reconstructions and digital phantom, with a scale bar of 2~mm. The heterogeneous SoS map follows a dual-region setting with a 1450~m/s background and a 1550~m/s ellipsoidal inclusion centered with semi-axes of 7, 8, and 11 mm. The detector array contains 4600 point transducers uniformly distributed on a spherical cap of radius 12.5 mm, with each detector recording 1024 samples at 50~MHz. The ground-truth initial pressure volume is generated at a voxel size of 0.1~mm and serves as the source for all simulations; it is not used directly as the quantitative evaluation target.


For the physical phantom dataset, the phantom contains absorptive structures and a tissue region with approximately 5\%--8\% acoustic contrast to the water. Measurements are acquired using a hemispherical array comprising 575 transducer elements. Data are collected at eight rotational views, yielding 4600 effective transducer positions in total. Each channel records 4096 samples at 20~MHz.

\subsubsection{Baselines}
We compare PAGS with three baselines that isolate different assumptions about PA source representation and SoS knowledge. All methods share identical data preprocessing steps.

\noindent\textbf{Vanilla UBP.}
This baseline applies standard universal back-projection with a globally uniform SoS of 1450~m/s.

\noindent\textbf{Dual-SoS UBP.}
This oracle variant assumes a known tissue--water boundary and performs back-projection with the corresponding path-averaged SoS, without modeling refraction. It serves as a prior-informed analytical reference for SoS compensation.

\noindent\textbf{Vanilla SlingBAG.}
This baseline follows the original SlingBAG algorithm~\cite{li2024slingbag}, using 10 spheres to approximate each Gaussian PA source and assuming a globally uniform SoS during forward projection.

\subsubsection{Evaluation}
For simulated data, we report peak signal-to-noise ratio (PSNR), root mean square error (RMSE), and structural similarity index measure (SSIM). Direct voxel-wise comparison against the original binary phantom volume is confounded by the absence of realistic opto-acoustic and transducer responses and by the binary, high-contrast nature of the phantom (values restricted to 0 and 1). To better isolate the ability to correct SoS heterogeneity, we construct a reference volume by simulating acoustic propagation from the same vascular phantom under a uniform 1450~m/s SoS using k-Wave, reconstructing with standard UBP, and peak-normalizing the result. All reconstructed volumes are peak-normalized in the same way. Because SoS mismatch can introduce a global geometric shift, we register each reconstruction to the reference via a rigid transformation (rotation and translation) estimated by maximizing normalized cross-correlation. This alignment corrects bulk positional errors without distorting local structures. The three metrics are then computed on these peak-normalized, rigidly registered volumes over a central region of interest spanning \(x\in[-2.5,5]\,\text{mm}\), \(y\in[-5,5]\,\text{mm}\), \(z\in[-10,5]\,\text{mm}\) with an isotropic voxel size of 0.1~mm. For physical phantom data, where such a reference is unavailable, we evaluate reconstruction quality by visual inspection, local contrast, and line-profile widths of fine structures.

\begin{figure*}[t]
    \centering
    \includegraphics[width=\linewidth]{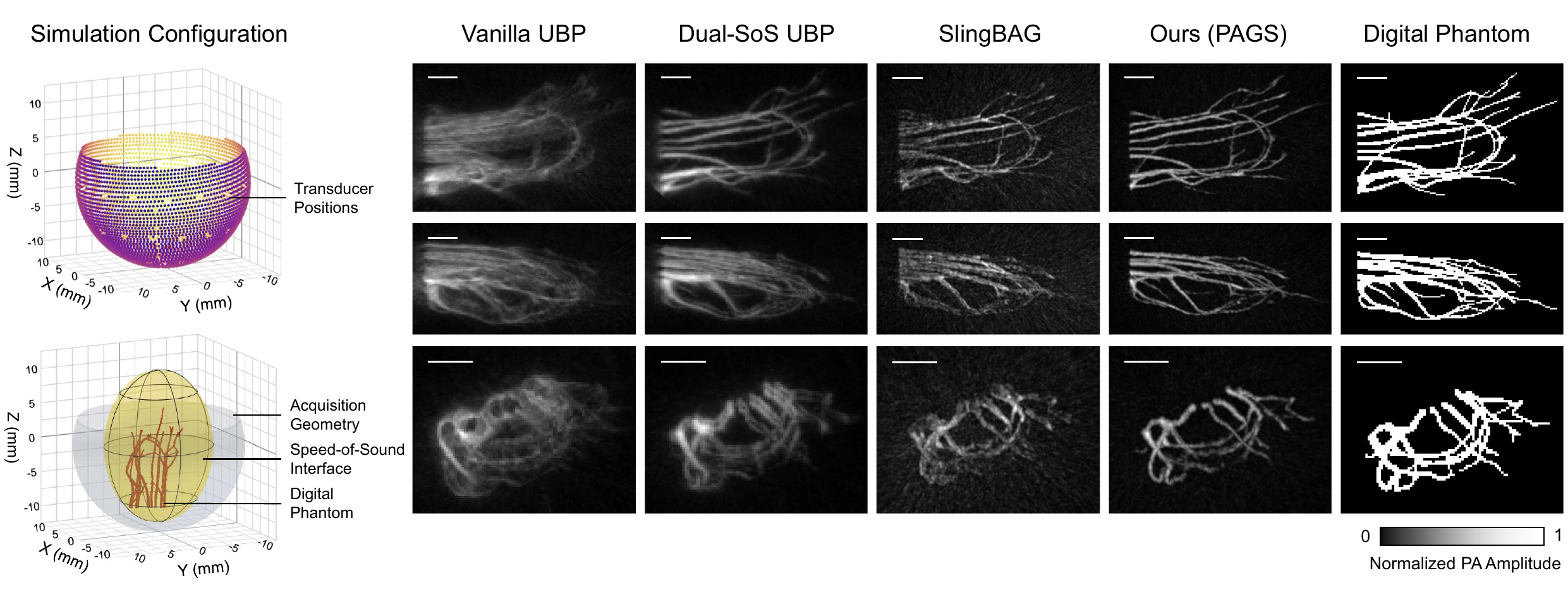}
    \caption{\textit{In silico} configuration and results. Left: transducer positions (top) and acquisition geometry with a grey spherical cap, a yellow ellipsoid speed-of-sound interface, and red phantom voxels of 0.1~mm edge length (bottom). Middle: qualitative reconstruction comparison in three orthogonal views. Right: corresponding digital phantom. Scale bar: 2~mm. PAGS corrects artifacts induced by a heterogeneous acoustic medium without calibrated SoS priors, yielding sharper vascular structures than uniform-SoS baselines.}
    \label{fig:insilico_results}
\end{figure*}

\subsubsection{Implementation}
PAGS is implemented in PyTorch. All experiments run on a graphics processing unit (GPU) workstation equipped with a single NVIDIA A100 (80~GB). The SH probe grid resolution is set between $8^3$ and $16^3$ depending on the physical size of the imaging scene, with the SH degree fixed at $D = 2$ (9 SH coefficients). Gaussian PA sources are initialized as approximately $10^5$--$10^6$ primitives with random initial center positions. Optimization uses the Adam optimizer ($\beta_1=0.9$, $\beta_2=0.999$); the learning rate for Gaussian positions is $10^{-4}$, for amplitudes and scales around $10^{-3}$, and for SH coefficients $10^{-2}$. Training proceeds for 200--500 iterations. Adaptive density control is performed every 50 iterations, splitting highly contributing primitives and pruning those with negligible contribution based on their signal reconstruction impact.

\subsection{In Silico Evaluation}
\label{subsec:exp_insilico}

We first evaluate PAGS on the \textit{in silico} dataset. This controlled setting tests whether PAGS can correct reconstruction errors induced by a heterogeneous acoustic medium without using the tissue--water boundary as a calibrated prior. Fig.~\ref{fig:insilico_results} compares the reconstructed vascular structures. Vanilla UBP suffers from visible geometric shift and defocus because a single uniform SoS cannot align the arrival times from different propagation paths. Vanilla SlingBAG benefits from a sparse Gaussian PA source representation and produces a cleaner background, but its uniform-SoS forward model still leaves noticeable blur and structural distortion. Dual-SoS UBP reduces part of the arrival-time mismatch by using the known tissue--water boundary, but it has two limitations. First, for efficient delay computation, the acoustic boundary must be represented by a simple geometry with tractable path-length calculation, such as an ellipsoid, which limits its ability to model more complex heterogeneous media. Second, it is an open-loop analytical reconstruction: once the path-averaged delays are computed, the fixed back-projection operator cannot adapt to residual signal mismatch.

In contrast, PAGS reconstructs sharper and more continuous vessel branches while maintaining a clean background. This improvement comes from optimizing the Gaussian PA sources jointly with the ASoS field, so the signal residual can update both the Gaussian PA source parameters and the path-level propagation correction.

Table~\ref{tab:insilico_metrics} reports PSNR, RMSE, and SSIM with respect to the uniform-SoS UBP reference. PAGS achieves the best scores among all compared methods, improving PSNR by 1.3~dB, reducing RMSE from 0.0354 to 0.0305, and increasing SSIM by 0.206 over vanilla SlingBAG. The substantial SSIM gain (from 0.331 to 0.537) indicates improved structural agreement rather than merely a visual sharpening effect.

\begin{table}[h]
    \centering
    \caption{Quantitative evaluation on the \textit{in silico} dataset, calculated against a uniform-SoS UBP reference after rigid registration and peak-intensity normalization.}
    \label{tab:insilico_metrics}
    \begin{tabular}{lccc}
        \toprule
        Method & PSNR $\uparrow$ & RMSE $\downarrow$ & SSIM $\uparrow$ \\
        \midrule
        Vanilla UBP & 23.2 & 0.0695 & 0.230 \\
        Dual-SoS UBP (Oracle) & 28.2 & 0.0388 & 0.384 \\
        Vanilla SlingBAG & 29.0 & 0.0354 & 0.331 \\
        \midrule
        \textbf{PAGS (Ours)} & \textbf{30.3} & \textbf{0.0305} & \textbf{0.537} \\
        \bottomrule
    \end{tabular}
\end{table}

\begin{figure*}[t]
    \centering
    \includegraphics[width=\linewidth]{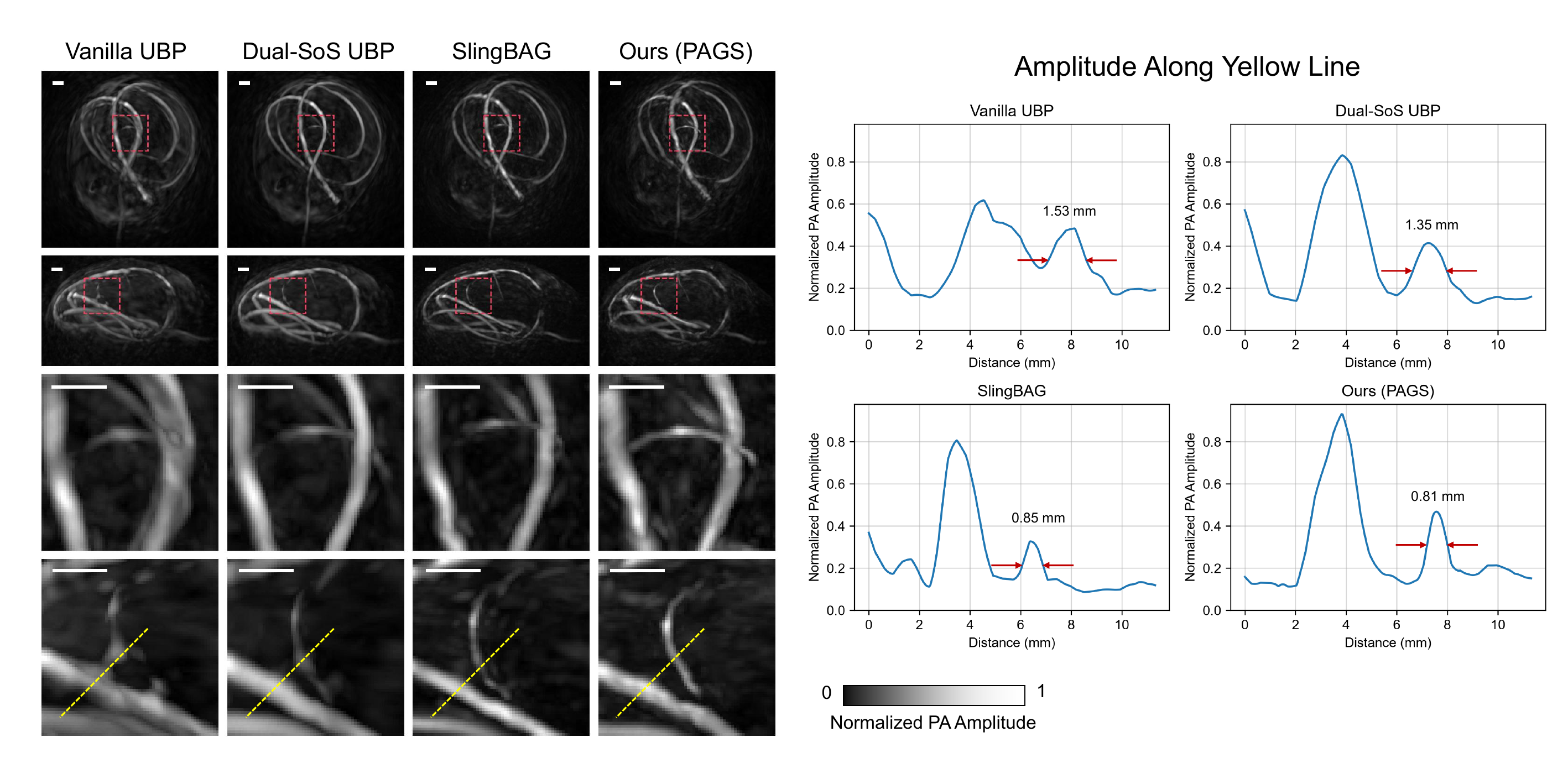}
    \caption{Physical phantom reconstruction results, scale bar: 5~mm. The left panels compare global views, zoomed regions marked by red boxes, and local line-profile locations marked in yellow. The right panels show the normalized PA amplitude profiles along the marked lines. Compared with uniform-SoS UBP, prior-informed Dual-SoS UBP, and vanilla SlingBAG, PAGS produces sharper absorber boundaries, improved local contrast, and clearer separation between neighboring structures under blind SoS adaptation.}
    \label{fig:physical_results}
\end{figure*}

\subsection{Experimental Validation on a Physical Phantom}
\label{subsec:exp_physical}

The \textit{in silico} experiment validates PAGS under known geometry and a controlled reference reconstruction. We next evaluate whether the same blind autofocusing mechanism remains effective on physical phantom measurements, where the recorded signals include real system effects such as finite transducer bandwidth, electronic noise, acoustic attenuation, and element-dependent sensitivity. Since voxel-level ground truth is unavailable, Fig.~\ref{fig:physical_results} focuses on visual structure and local profile sharpness.

The global views in Fig.~\ref{fig:physical_results} show that vanilla UBP suffers from broad structural blur and reduced contrast, consistent with uncorrected SoS mismatch. Dual-SoS UBP improves part of the alignment by using a prescribed tissue--water boundary, but its simple boundary model and open-loop back-projection still leave residual defocus. Vanilla SlingBAG suppresses part of the diffuse background through its sparse Gaussian PA source representation, yet its uniform-SoS forward model cannot fully focus the absorbers.

The zoomed regions and line-profile comparisons provide a more local view of this effect. The higher-contrast absorber responses indicate improved focusing rather than only a change in global intensity scaling. Neighboring structures that appear blurred or partially merged in the uniform-SoS baselines become more separable after joint source and ASoS optimization.

Taken together, the image-domain sharpening and improved local contrast suggest that PAGS can compensate for dominant arrival-time errors in real measurements while remaining a blind reconstruction method.

\subsection{Robustness under Sparse Sampling}
\label{subsec:exp_sparse}

We further evaluate PAGS under sparse-view acquisition by uniformly retaining 50\%, 25\%, and 12.5\% of the detector channels from the physical phantom measurements. This experiment tests whether the proposed representation remains stable when the inverse problem becomes increasingly underdetermined.

\begin{figure}[htbp]
    \centering
    \includegraphics[width=\linewidth]{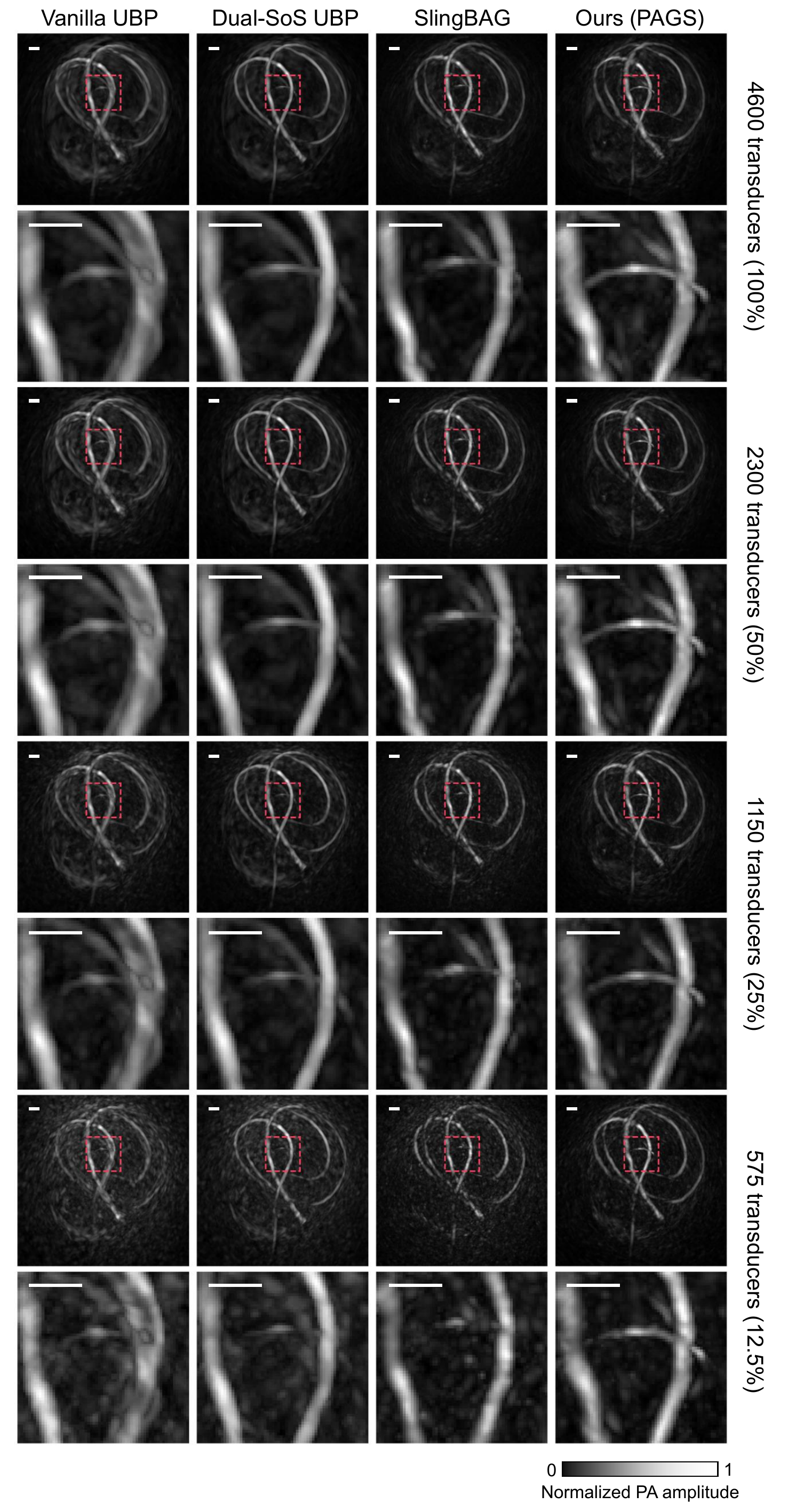}
    \caption{Robustness under sparse-view sampling on the physical phantom data, scale bar: 5~mm. PAGS is evaluated by retaining 50\%, 25\%, and 12.5\% of the detector channels. The reconstruction degrades gradually as sampling becomes sparser, while major vascular structures and branch connectivity remain visible.}
    \label{fig:sparse_results}
\end{figure}

Fig.~\ref{fig:sparse_results} shows that PAGS degrades gracefully as the sampling ratio decreases. Major vascular trunks and branch connectivity remain visible even at 12.5\% sampling, while the main degradation appears as reduced fine-detail contrast rather than severe streak artifacts. This behavior suggests that the Gaussian PA source representation provides an effective sparse structural constraint under limited measurements.

The robustness also benefits from the propagation model.
The low-order SH parameterization biases the ASoS field toward smooth path-level variations, reducing the risk of fitting sparse-view artifacts with nonphysical high-frequency SoS changes.
As a result, PAGS maintains a compact source distribution and a stable propagation correction even when fewer detector channels are available.
Rotating 3D views of the full-method comparison and the sparse-view results are provided in Supplementary Video~1.

\subsection{Ablation Study and Efficiency Analysis}
\label{subsec:exp_ablation}

We isolate the two main technical components of PAGS: the analytic Gaussian acoustic projection and the SH-parameterized ASoS field.

\noindent\textbf{Ablation design.}
We use a $2\times2$ design that combines either the 10-sphere or analytic forward model with either a uniform SoS or the ASoS field:
\begin{enumerate}
    \item \textbf{Baseline}: 10-sphere forward model with a uniform SoS, corresponding to vanilla SlingBAG.
    \item \textbf{+ ASoS field}: 10-sphere forward model with the SH-parameterized ASoS field.
    \item \textbf{+ Analytic}: analytic Gaussian forward model with a uniform SoS.
    \item \textbf{Full PAGS}: analytic Gaussian forward model with the SH-parameterized ASoS field.
\end{enumerate}
All variants use the same initialization, optimizer settings, random seed, and number of iterations. Adaptive density control is applied every 50 iterations, and the optimizer state is reset after each density-control step, which explains the transient spikes in the loss curves.

\begin{figure}[htbp]
    \centering
    \includegraphics[width=0.9\linewidth]{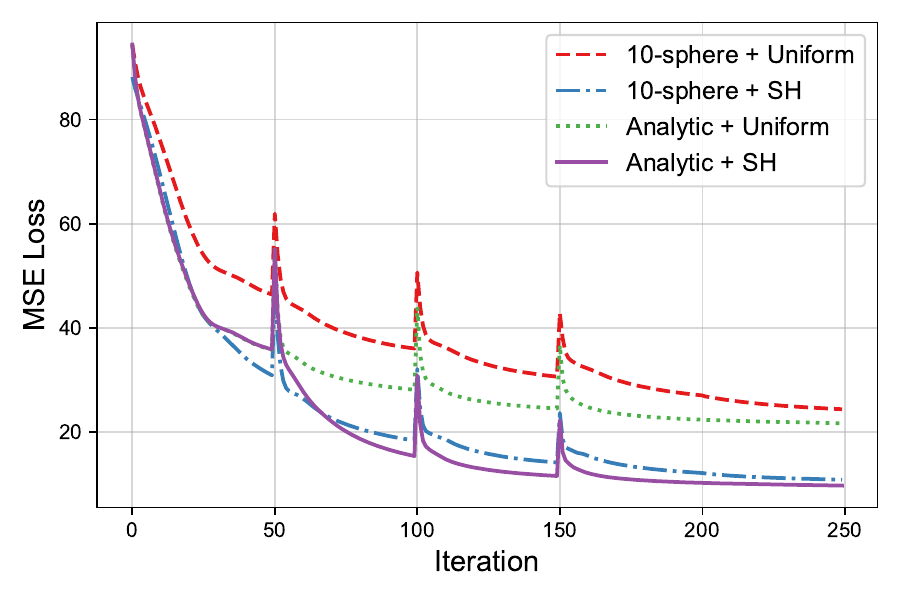}
    \caption{Mean squared error (MSE) loss convergence of the four ablation configurations on the heterogeneous-SoS simulation. The ASoS field substantially reduces the final signal residual, while the analytic Gaussian forward model lowers the computational cost of each optimization iteration.}
    \label{fig:ablation_loss}
\end{figure}

Fig.~\ref{fig:ablation_loss} compares the optimization behavior of the four configurations. Introducing the ASoS field substantially lowers the final signal-domain residual, indicating that the latent propagation correction provides additional flexibility for matching measurements acquired through a heterogeneous acoustic medium. Replacing the 10-sphere approximation with the analytic Gaussian model yields a similar convergence trend with markedly lower per-iteration cost. Full PAGS combines both components and reaches the lowest final signal residual.

Table~\ref{tab:efficiency} reports the per-iteration forward--backward runtime under identical hardware and problem scale. The analytic Gaussian forward model reduces runtime from 6.7~s to 2.3~s by replacing repeated error-function evaluations in the 10-sphere approximation with a single closed-form Gaussian response. Adding the ASoS field introduces only minor overhead, increasing runtime from 2.3~s to 2.4~s in the analytic setting. This shows that the proposed propagation model improves focusing while preserving most of the computational benefit of the analytic projection.

\begin{table}[htbp]
    \centering
    \caption{Computational efficiency of the four ablation configurations. The analytic Gaussian forward model provides the main speedup, while adding the ASoS field introduces only minor overhead.}
    \label{tab:efficiency}
    \scriptsize
    \setlength{\tabcolsep}{2pt}
    \begin{tabular}{@{}p{0.25\linewidth}p{0.27\linewidth}p{0.18\linewidth}c@{}}
        \toprule
        Configuration & Forward model & SoS field & \shortstack{Time/iter. (s) $\downarrow$} \\
        \midrule
        Baseline & 10-sphere approx. & Uniform & 6.7 \\
        + ASoS field & 10-sphere approx. & SH probes & 6.9 \\
        + Analytic & Analytic & Uniform & 2.3 \\
        \textbf{Full PAGS} & \textbf{Analytic} & \textbf{SH probes} & \textbf{2.4} \\
        \bottomrule
    \end{tabular}
\end{table}

\section{Conclusion}
\label{sec:conclusion}

We presented PAGS, a differentiable autofocusing framework for PACT under unknown SoS heterogeneity. Instead of recovering a dense acoustic medium, PAGS learns an anisotropic path-averaged SoS field whose values directly determine source-to-transducer arrival-time alignment. This latent propagation model is coupled with sparse Gaussian PA sources and an analytic Gaussian acoustic projection, enabling signal-domain optimization of both source and propagation parameters without calibrated SoS priors. Experiments on simulated and physical phantom data show sharper reconstruction under heterogeneous acoustic media, robust behavior under sparse-view sampling, and complementary benefits from the ASoS field and the analytic projection. PAGS estimates an effective path-level propagation field rather than a physical SoS map, and the present validation is limited to simulated and phantom data. Future work will extend this path-level formulation to more complex acoustic settings and broader biological validation.

\section*{Abbreviations}
{\small
\renewcommand{\arraystretch}{1.05}
\setlength{\tabcolsep}{4pt}
\noindent
\begin{tabular}{@{}>{\RaggedRight\arraybackslash}p{0.19\linewidth}>{\RaggedRight\arraybackslash}p{0.75\linewidth}@{}}
\toprule
\textbf{Abbrev.} & \textbf{Definition} \\
\midrule
3DGS & 3D Gaussian splatting \\
ASoS & Anisotropic path-averaged speed of sound \\
CT & Computed tomography \\
DAS & Delay-and-sum \\
GPU & Graphics processing unit \\
MSE & Mean squared error \\
PA & Photoacoustic \\
PACT & Photoacoustic computed tomography \\
PAGS & Speed-of-sound-adaptive Gaussian splatting \\
PSNR & Peak signal-to-noise ratio \\
RMSE & Root mean square error \\
SH & Spherical harmonics \\
SoS & Speed of sound \\
SSIM & Structural similarity index measure \\
ToF & Time of flight \\
TV & Total variation \\
UBP & Universal back-projection \\
\bottomrule
\end{tabular}
}

\section*{Declarations}

\subsection*{Data Availability}
The datasets generated and analyzed during the current study are available from the corresponding author on reasonable request.

\subsection*{Code Availability}
The official implementation of PAGS, together with usage instructions and a synthetic example, is publicly available at \url{https://github.com/work-submit/PAGS/}.

\subsection*{Competing Interests}
The authors have no competing interests to declare that are relevant to the content of this article.

\subsection*{Author Contributions}
Jiarui Ge developed the methodology and algorithms, implemented the software, conducted the computational experiments, and analyzed and visualized the results. Jintao Ma, Bangxu Fan, and Jinyan Zhang performed the photoacoustic data simulations and experimental data acquisition. Xiaokang Yang provided overall supervision and strategic scientific guidance. Shuai Na and Xiaoyun Yuan jointly supervised the work, participated in regular project discussions, advised on technical details, and guided manuscript preparation and revision. All authors discussed the results, contributed to writing and revising the manuscript, and approved the final version.

\subsection*{Funding}
This work was supported by the National Natural Science Foundation of China (NSFC) under Grant 62271283, the National Key Research and Development Program of China under Grant 2024YFC2421300, the Jiangsu Provincial Frontier Technology R\&D Program under Grant BF2025002, and the Shanghai Jiao Tong University Medical and Engineering Cross Research Fund under Grant YG2026QNB50.

\subsection*{Acknowledgements}
Not applicable.

\bibliographystyle{unsrt}
\bibliography{reference}

\end{document}